\documentclass[runningheads]{llncs}
\usepackage{graphicx}
\usepackage{amsmath,amssymb} 
\usepackage{epsfig}
\usepackage{color}
\usepackage{booktabs} 
\usepackage{multirow}
\usepackage{subcaption}
\usepackage[font={footnotesize}]{caption}
\usepackage[pagebackref=true,breaklinks=true,letterpaper=true,colorlinks,bookmarks=false]{hyperref}
\usepackage{xspace}
\usepackage{xcolor}

\usepackage[breaklinks=true,bookmarks=false]{hyperref}
\hypersetup{
    colorlinks,
    linkcolor={red!90!black},
    citecolor={green!90!black},
    urlcolor={blue!90!black}
}

\usepackage[english]{babel}

\begin{document}
\pagestyle{headings}
\mainmatter
\def\ECCVSubNumber{1600}  

\title{Renovating Parsing R-CNN for Accurate Multiple Human Parsing} 



\titlerunning{RP R-CNN}
%
\author{Lu Yang\inst{1}\orcidID{0000-0003-3857-3982} \and
Qing Song\inst{1} \thanks{The corresponding author is Qing Song.}\orcidID{0000-0003-4616-2200} \and
Zhihui Wang\inst{1}\orcidID{0000-0002-7547-8864} \and
Mengjie Hu\inst{1}\orcidID{0000-0001-7712-3322} \and
Chun Liu\inst{1}\orcidID{0000-0002-2834-9461} \and
Xueshi Xin\inst{1}\orcidID{0000-0002-9326-4499} \and
Wenhe Jia\inst{1}\orcidID{0000-0002-2516-957X} \and
Songcen Xu\inst{2}\orcidID{0000-0002-0022-0906}}
\authorrunning{L. Yang et al.}
%
\institute{Beijing University of Posts and Telecommunications Beijing 100876, China \and
Noah's Ark Lab,  Huawei Technologies\\
\email{\{soeaver, priv, wangzh, mengjie.hu, chun.liu, xinxueshi, srxhemailbox\}@bupt.edu.cn}\\
\email{xusongcen@huawei.com}}
\maketitle

\begin{abstract}
Multiple human parsing aims to segment various human parts and associate each part with the corresponding instance simultaneously. This is a very challenging task due to the diverse human appearance, semantic ambiguity of different body parts, and complex background. Through analysis of multiple human parsing task, we observe that human-centric global perception and accurate instance-level parsing scoring are crucial for obtaining high-quality results. But the most state-of-the-art methods have not paid enough attention to these issues. To reverse this phenomenon, we present Renovating Parsing R-CNN (RP R-CNN), which introduces a global semantic enhanced feature pyramid network and a parsing re-scoring network into the existing high-performance pipeline. The proposed RP R-CNN adopts global semantic representation to enhance multi-scale features for generating human parsing maps, and regresses a confidence score to represent its quality. Extensive experiments show that RP R-CNN performs favorably against state-of-the-art methods on CIHP and MHP-v2 datasets. Code and models are available at \url{https://github.com/soeaver/RP-R-CNN}.

\end{abstract}

\section{Introduction}
Multiple human parsing~\cite{Gong_eccv2018_pgn}~\cite{Li_arxiv2017_mhparser}~\cite{Yang_cvpr2019_parsingrcnn} is a fundamental task in multimedia and computer vision, which aims to segment various human parts and associate each part with the corresponding instance. It plays a crucial role in applications in human-centric analysis and potential down-stream applications, such as person re-identification~\cite{Li_cvpr2018_harmonious}~\cite{Miao_iccv2019_pose}, action recognition~\cite{Girdhar_nips2017_attpool}, human-object interaction~\cite{Chao_iccv2015_hico}~\cite{Qi_eccv2018_learning}~\cite{Gkioxari_cvpr2018_interacnet}, and virtual reality~\cite{Hsieh_mm2019_fashion}.

Due to the successful development of convolutional neural networks~\cite{Russakovsky_ijcv2015_imagenet}~\cite{He_cvpr2016_resnet}, great progress has been made in multiple human parsing. Current state-of-the-art methods can be categorized into bottom-up, one-stage top-down, and two-stage top-down methods. The bottom-up methods~\cite{Gong_eccv2018_pgn}~\cite{Gong_cvpr2019_graphonomy}~\cite{He_aaai2020_grapyml} regard multiple human parsing as a fine-grained semantic segmentation task, which predicts the category of each pixel and grouping them into corresponding human instance. This series of methods will have better performance in semantic segmentation metrics, but poor in instance parsing metrics, especially easy to confuse adjacent human instances. Unlike bottom-up methods, the one-stage top-down~\cite{Yang_cvpr2019_parsingrcnn}~\cite{Qin_bmvc2019_unified} and two-stage top-down methods~\cite{Ruan_aaai2019_ce2p}~\cite{Liu_mm2019_braidnet}~\cite{Ji_arxiv2019_sematree} locate each instance in the image plane, and then segment each human parts independently. The difference between one-stage and two-stage is whether the detector is trained together with the sub-network used to segment the human part in an end-to-end manner. Compared with the bottom-up, the top-down methods are very flexible, which can easily introduce enhancement modules or train with other human analysis tasks (such as pose estimation~\cite{Xiao_eccv2018_simple}, dense pose estimation~\cite{Guler_cvpr2018_densepose}~\cite{Yang_cvpr2019_parsingrcnn} or clothing parsing~\cite{Pongsate_pami2014_retrieving}) jointly. So it has become the mainstream research direction of multiple human parsing. But the human parts segmentation of each instance is independent and cannot make full use of context information, so the segmentation of some small scale human parts and human contours still needs to be improved. In addition, it is worth noting that neither bottom-up or top-down methods have a good way to evaluate the quality of predicted instance parsing maps. Resulting in many low-quality results that cannot be filtered.

\begin{figure*}[t]
\begin{center}
\includegraphics[width=0.90\linewidth]{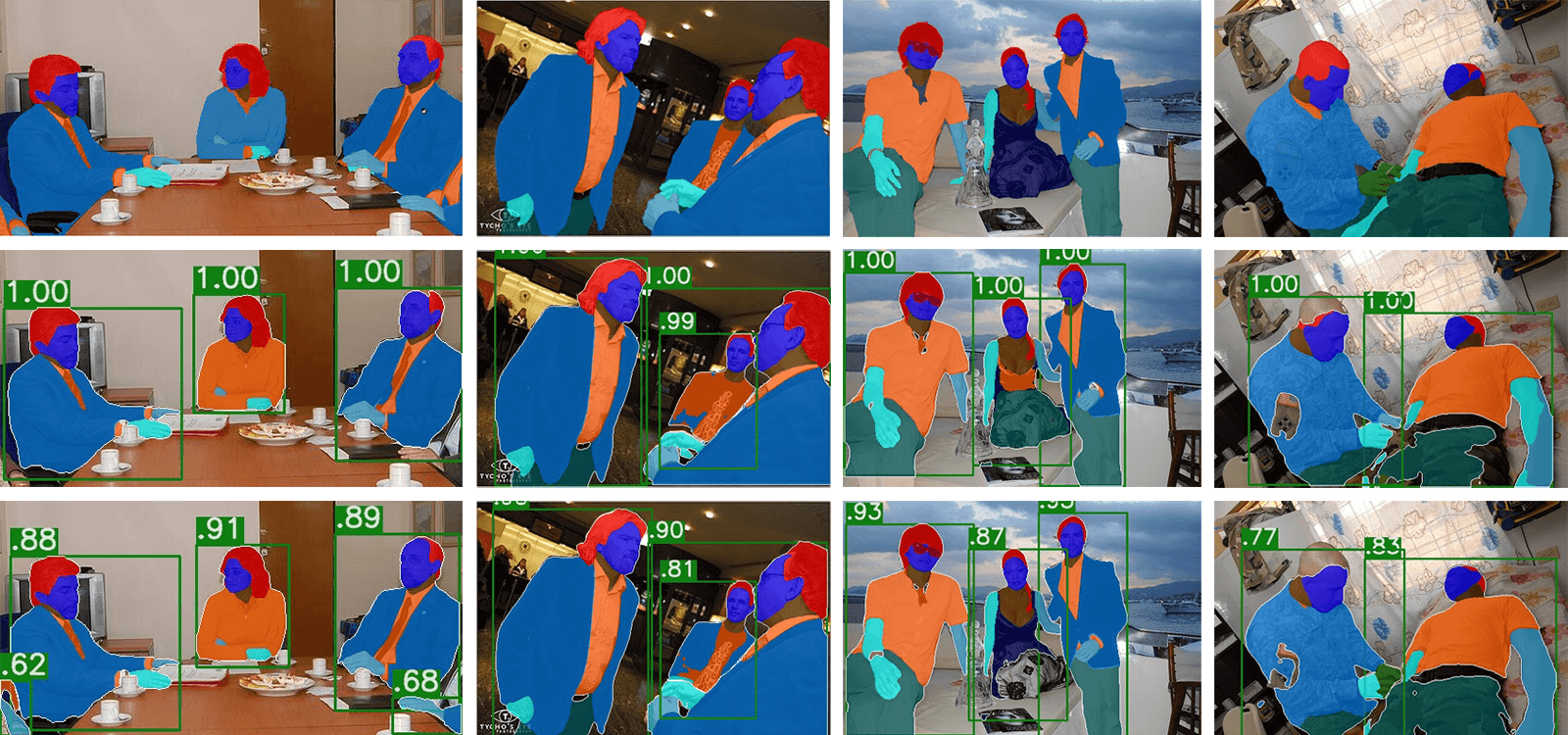}
\end{center}
\caption{Comparison of results between Parsing R-CNN and RP R-CNN on CIHP dataset. The first row is the ground-truth, the second row is the results of Parsing R-CNN, and the third is the predictions of RP R-CNN.}
\label{fig:intro}
\end{figure*}

In this paper, we are devoted to solving the problem of missing global semantic information in top-down methods, and evaluating the quality of predicted instance parsing maps accurately. Therefore, we propose Renovating Parsing R-CNN (RP R-CNN), which introduces a global semantic enhanced feature pyramid network and a parsing re-scoring network to renovate the pipeline of top-down multiple human parsing. The global semantic enhanced feature pyramid network (GSE-FPN) is built on the widely used FPN~\cite{Lin_cvpr2017_fpn}. We up-sample the multi-scale features generated by FPN to the same scale and fuse them. Using the global human parts segmentation to supervise and generate the global semantic feature, then fusing the global semantic feature with FPN features on the corresponding scales. GSE-FPN encourages the semantic supervision signal to directly propagate to the feature pyramid, so as to strengthen global information of learned multi-scale features. Global semantic enhanced features are passed to the Parsing branch through the RoIAlign~\cite{He_iccv2017_maskrcnn} operation, ensuring that each independent human instance can still perceive the global semantic information, thereby improving the parsing performance of small targets, human contours, and easily confused categories. On this basis, the parsing re-scoring network (PRSN) is used to sense the quality of instance parsing maps and gives accurate scores. The score of instance parsing map is related to filtering low-quality results and sorting of instances, which is very important in the measurement of method and practical application. However, almost all the top-down methods use the score of detected bounding-box to represents the quality of instance parsing map~\cite{Ruan_aaai2019_ce2p}~\cite{Yang_cvpr2019_parsingrcnn}. This will inevitably bring great deviation, because the score of bounding-box can only indicate whether the instance is human or not, while the score of the instance parsing map needs to express the segmentation accuracy of each human part, and there is no direct correlation between them. The proposed PRSN is a very lightweight network, taking the feature map and heat map of each human instance as input, using MSE loss to regress the mean intersection over union (mIoU) between the prediction and ground-truth. During inference, we use the arithmetic square root of predicted mIoU score multiply box classification score as human parsing final score. 

Extensive experiments are conducted on two challenging benchmarks, CIHP~\cite{Gong_eccv2018_pgn} and MHP-v2~\cite{Zhao_mm2018_mhpv2}, demonstrating that our proposal RP R-CNN significantly outperforms the state-of-the-art for both bottom-up and top-down methods. As shown is Figure~\ref{fig:intro}, RP R-CNN is more accurate in segmenting small parts and human edges, and the predicted parsing scores can better reflect the quality of instance parsing maps. The main contributions of this work are summarized as follows:

\begin{itemize}
\item
A novel RP R-CNN is proposed to solve the issue of missing global semantic information and inaccurate scoring of instance parsing maps in top-down multiple human parsing.

\item
We introduce an effective method to improve the multiple human parsing results by fusing global and instance-level human parts segmentation.

\item
The proposed RP R-CNN achieves state-of-the-art on two challenging benchmarks. On CIHP \texttt{val} set, RP R-CNN yields 2.0 points mIoU and 7.0 points AP$^\text{p}_\text{50}$ improvements compared with Parsing R-CNN~\cite{Yang_cvpr2019_parsingrcnn}. On MHP-v2 \texttt{val} set, RP R-CNN outperforms Parsing R-CNN by 13.9 points AP$^\text{p}_\text{50}$ and outperforms CE2P~\cite{Ruan_aaai2019_ce2p} by 6.0 points AP$^\text{p}_\text{50}$, respectively.
\end{itemize}
Our code and models of RP R-CNN are publicly available.

\section{Related Work}

\noindent\textbf{Multi-Scale Feature Representations.} Multi-scale feature is widely used in computer vision tasks~\cite{Lin_cvpr2017_fpn}~\cite{Kirillov_cvpr2019_pfpn}~\cite{Long_cvpr2015_fcn}~\cite{Xiao_eccv2018_upernet}. Long {\em et al.}~\cite{Long_cvpr2015_fcn} combine coarse, high layer information with fine, low layer information to generate fine features with high resolution, which greatly promotes the development of semantic segmentation. Lin {\em et al.}~\cite{Long_cvpr2015_fcn} present the feature pyramid network (FPN), and adopt it in object detection, greatly improve the performance of the small object. FPN is a feature pyramid with high-level semantics throughout, through top-down pathway and lateral connections. With the success of FPN, some researches introduce it into other tasks. Panoptic feature pyramid network (PFPN)~\cite{Kirillov_cvpr2019_pfpn} is proposed by Kirillov {\em et al.} and applied to panoptic segmentation. PFPN up-samples the feature pyramids and fuse them to the same spatial resolution, then a semantic segmentation branch is attached to generate high-resolution semantic features. However, the computation cost of PFPN is too large, and it only has a single scale semantic feature. Our GSE-FPN solves the above problems well, which adopts a lightweight up-sampling method, and use the global semantic feature to enhance the multi-scale feature.

\noindent\textbf{Instance Scoring.} Scoring the predicted instance is a challenging question. The R-CNN series~\cite{Girshick_iccv2015_fast-rcnn}~\cite{Ren_nips2015_faster-rcnn} of object detection approaches use the object classification score as the confidence of detection results. Recent studies~\cite{Jiang_eccv2018_iounet}~\cite{Tan_iccv2019_ltr}~\cite{Zhu_arxiv2020_cpm} believe that it cannot accurately reflect the consistency between the predicted bounding-box and the ground-truth. Jiang {\em et al.}~\cite{Jiang_eccv2018_iounet} present the IoU-Net, which adopts a IoU-prediction branch to predict the IoU between the predicted bounding box and the corresponding ground truth. Tan {\em et al.}~\cite{Tan_iccv2019_ltr} propose the Learning-to-Rank (LTR) model to produce a ranking score, which is based on IoU to indicate the ranks of candidates during the NMS step. Huang {\em et al.}~\cite{Huang_cvpr2019_msrcnn} consider the difference between classification score and mask quality is greater in instance segmentation. They proposed Mask Scoring R-CNN, which uses a MaskIoU head to predict the quality of mask result. Different from these studies, this work analyzes and solves the inaccurate scoring of instance parsing maps for the first time. The proposed PRSN is concise yet effective, and reducing the gap between score and instance parsing quality.

\noindent\textbf{Multiple Human Parsing.} Before the popularity of convolutional neural network, some methods~\cite{Yamaguchi_cvpr2012_parsing}~\cite{Luo_iccv2013_pedestrian}~\cite{Yang_cvpr2014_clothing-co-parsing} using hand-crafted visual features and low-level image decompositions have achieved considerable results on single human parsing. However, limited by the representation ability of features, these traditional methods can not be well extended to multiple human parsing. With the successful development of convolutional neural networks~\cite{Russakovsky_ijcv2015_imagenet}~\cite{Alex_nip2012_alexnet}~\cite{He_cvpr2016_resnet} and open source of large-scale multiple human parsing datasets~\cite{Gong_eccv2018_pgn}~\cite{Zhao_mm2018_mhpv2}, some recent researches~\cite{Gong_eccv2018_pgn}~\cite{Ruan_aaai2019_ce2p}~\cite{Yang_cvpr2019_parsingrcnn} have achieved remarkable results in instance-level multiple human parsing. Gong {\em et al.}~\cite{Gong_eccv2018_pgn} present the part grouping network (PGN), which is a typical bottom-up method for instance-level multiple human parsing. PGN reformulates multiple human parsing as semantic part segmentation task and instance-aware edge detection task, the former is used to assign each pixel as human part and the latter is used to group semantic part into different human instances. Ruan {\em et al.}~\cite{Ruan_aaai2019_ce2p} rethink and analyze the problems of feature resolution, global context information and edge details in human parsing task, and propose Context Embedding with Edge Perceiving (CE2P) framework for single human parsing. CE2P is a very successful two-stage top-down method, and wins the 1st places on three tracks in the 2018 2nd LIP Challenge. Parsing R-CNN~\cite{Yang_cvpr2019_parsingrcnn} is proposed by Yang {\em et al.}, which is a one-stage top-down method for multiple human parsing. Based on the in-depth analysis of human appearance characteristics, Parsing R-CNN has made an effective extension on region-based approaches~\cite{Girshick_iccv2015_fast-rcnn}~\cite{Ren_nips2015_faster-rcnn}~\cite{Lin_cvpr2017_fpn}~\cite{He_iccv2017_maskrcnn} and significantly improved the performance of human parsing. Our work is based on the Parsing R-CNN framework, firstly introducing the global semantic information and reducing the gap between score and instance parsing quality for the top-down methods.

\section{Renovating Parsing R-CNN}

Our goal is to solve the issue of missing global semantic information and inaccurate scoring of instance parsing map in top-down multiple human parsing pipeline. In this section, we will introduce the motivation, architecture, and components of RP R-CNN in detail.

\begin{table*}[t]
\centering
\small
\tabcolsep 0.04in 
\scalebox{0.88}{
\begin{tabular}{c|ccc|cccc}
Backbones &  GT-box &  GT-parsing & GT-score & mIoU  &  AP$^\text{p}_\text{50}$ & AP$^\text{p}_\text{vol}$ & PCP$_\text{50}$                     \\
  \hline  
 \multirow{4}{*}{R50-FPN}   &                  &  &   & 56.2 & 64.6 & 54.3 & 60.9    \\
                                            \cline{5-8}
                                            & \checkmark &  & & ${58.4}_{\color{red}(+2.2)}$ & ${58.0}_{\color{blue}(-6.6)}$ & ${51.5}_{\color{blue}(-2.8)}$ & ${62.3}_{\color{red}(+1.4)}$  \\ 
                                            &  &\checkmark&   & ${87.8}_{\color{red}(+31.6)}$ & ${91.4}_{\color{red}(+26.8)}$ & ${83.6}_{\color{red}(+29.3)}$ & ${90.6}_{\color{red}(+29.7)}$  \\ 
                                            & & &\checkmark   & ${57.4}_{\color{red}(+1.2)}$ & ${73.7}_{\color{red}(+9.1)}$ & ${60.8}_{\color{red}(+6.5)}$ & ${60.9}_{(+0.0)}$   \\
\end{tabular}
}
  \caption{Upper bound analysis of instance-level multiple human parsing via using ground-truth. All models are trained on CIHP \texttt{train} set and evaluated on CIHP \texttt{val} set. We replace the Bbox branch output with ground-truth box, replace the Parsing branch output with ground-truth segmentation or replace the instance score with ground-truth IoU, respectively. The results suggest that there is still room for improvement in human parsing and scoring.}
  \label{tab:upper_bound}
\end{table*}

\subsection{Motivation}

In the top-down multiple human parsing pipeline, the network outputs three results: bounding-box, instance parsing map and parsing score. The importance of three outputs to the network performance is different. We take Parsing R-CNN~\cite{Yang_cvpr2019_parsingrcnn} as baseline, and make an upper bound analysis of the three outputs. As shown in Table~\ref{tab:upper_bound}, we replace the predicted bounding-box with ground-truth, the multiple human parsing increases by 2.2 points mIoU~\cite{Long_cvpr2015_fcn}, but AP$^\text{p}_\text{50}$ and  AP$^\text{p}_\text{vol}$~\cite{Zhao_mm2018_mhpv2} decrease. But when we replace the corresponding network output with the ground-truth of parsing map and mIoU, all the evaluation metrics have significant improvements. In particular, after the adoption of ground-truth parsing map, each evaluation metric has increased by about 30 points. These experimental results show that the accuracy of bounding-box has no significant impact on the multiple human parsing performance.However, the predicted parsing map and score still have a lot of room for improvement and not been paid enough attention by current studies. This is the motivation for our work.

\begin{figure*}[t]
\begin{center}
\includegraphics[width=0.98\linewidth]{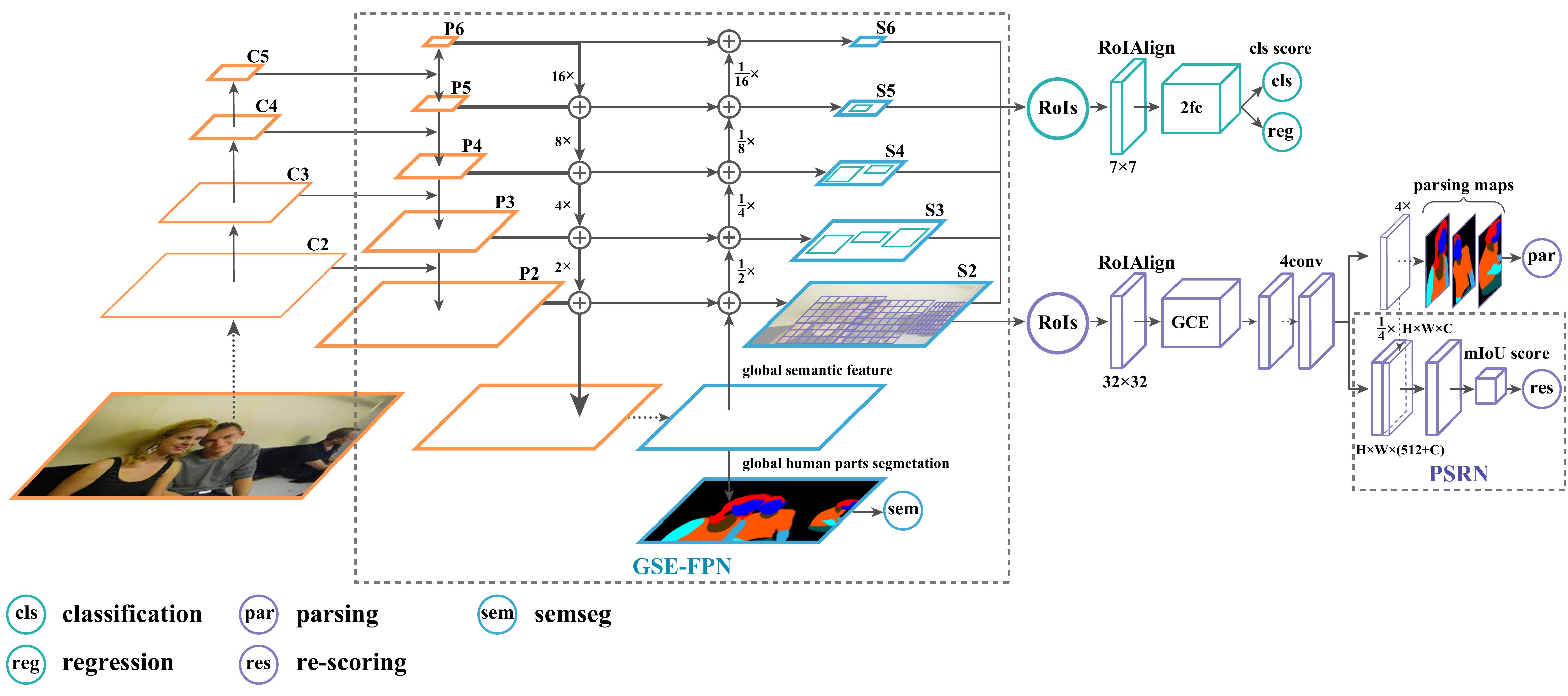}
\end{center}
\caption{RP R-CNN architecture. The input image is fed into a backbone with Global Semantic Enhanced FPN~\cite{Lin_cvpr2017_fpn} to generate RoIs via RPN~\cite{Ren_nips2015_faster-rcnn} (not shown in the figure) and RoI features via RoIAlign~\cite{He_iccv2017_maskrcnn}. The global human parts segmentation is used to supervise and generate the global semantic feature. The BBox branch is standard component of Faster R-CNN which is used to detect human instance. The Parsing branch is mainly composed of GCE module~\cite{Yang_cvpr2019_parsingrcnn} and Parsing Re-Scoring Network for predicting parsing maps and mIoU scores.}
\label{fig:rp_rcnn_arch}
\end{figure*}

\subsection{Architecture}

As illustrated in Figure~\ref{fig:rp_rcnn_arch}, the proposed RP R-CNN involves four components: Backbone, GSE-FPN, Detector (RPN and BBox branch), and Parsing branch with PRSN. The settings of Backbone and Detector are the same as Parsing R-CNN~\cite{Yang_cvpr2019_parsingrcnn}. The GSE-FPN is attached to the Backbone to generate multi-scale features with global semantic information. The Parsing branch consists of GCE module, parsing map output and Parsing Re-Scoring Network.

\subsection{Global Semantic Enhanced Feature Pyramid Network}

\begin{figure*}[t]
\begin{center}
\includegraphics[width=0.70\linewidth]{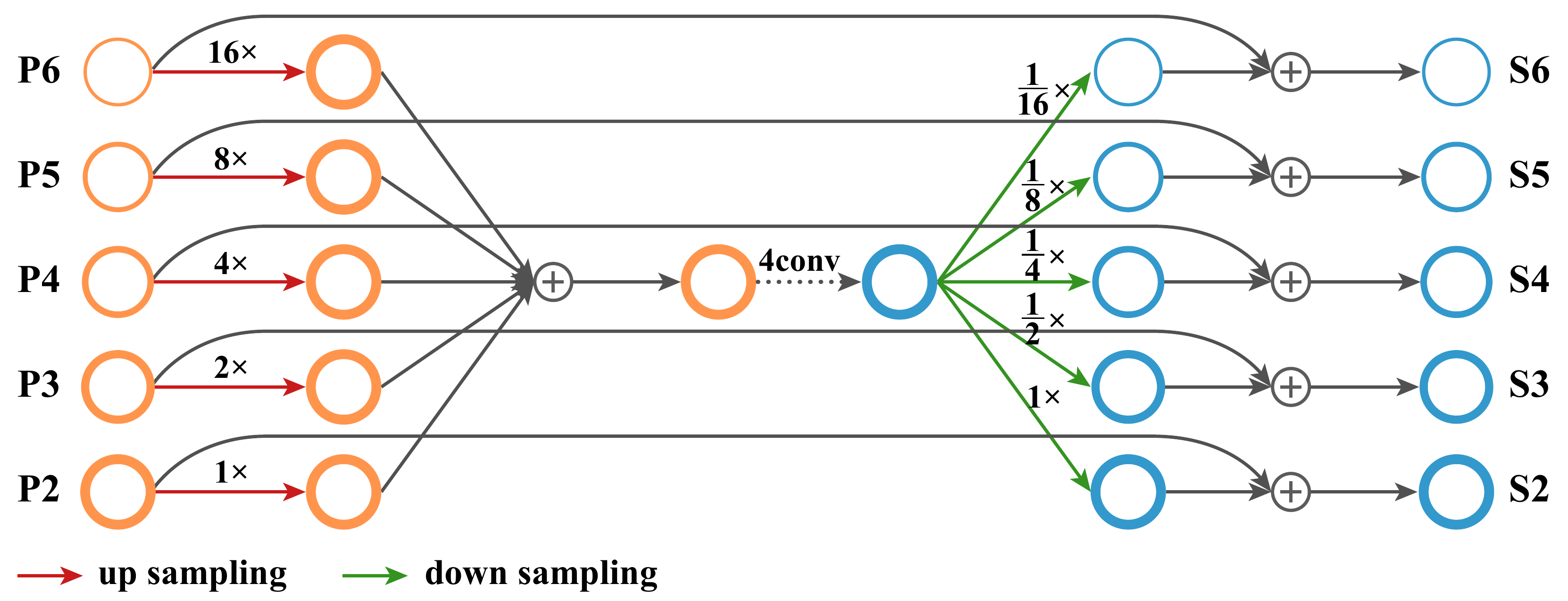}
\end{center}
\caption{Global semantic enhanced feature pyramid network (GSE-FPN). Circle is used to represent the feature map, and the circle thickness of circle is used to represent the spatial scale. The semantic segmentation loss is omitted here.}
\label{fig:sfpn}
\end{figure*}

RoIAlign~\cite{He_iccv2017_maskrcnn} aims to obtain the features of a specific region on the feature map, so that each instance can be processed separately. However, this makes the instance unable to directly perceive the global (context) information in branch. Global representation is crucial for human parsing, because we not only need to distinguish human body and background, but also give each pixel corresponding category through understanding the pose and recognizing the clothes the person wears~\cite{Gong_cvpr2019_graphonomy}. Therefore, the information about the environment and objects around the human body is helpful for network learning. Some methods~\cite{Zeng_pami2017_gbd} perceive more valuable information by changing the area selected by RoIPool/RoIAlign. Different from these, we hope that by explicitly enhancing the global semantic representation of multi-scale features before RoIAlign. As a concrete example, the proposed GSE-FPN are illustrated in Figure~\ref{fig:sfpn}, we adopt group normalization~\cite{Wu_eccv2018_gn} and ReLU activation~\cite{Nair_icml2010_relu} after each convolutional layer.

\noindent\textbf{High-resolution Feature.} For semantic segmentation, high-resolution feature is necessary for generating high-quality results. Dilated convolution is an effective operation, which is adopted by many state-of-the-art semantic segmentation methods~\cite{Yu_iclr2016_dilated}~\cite{Zhao_cvpr2017_pspnet}~\cite{Chen_arxiv2017_deeplabv3}. But dilated convolution substantially increases computation cost, and limits the use of multi-scale features. To keep the efficiency of network, and generate high-resolution features, we extend the multi-scale outputs of FPN~\cite{Lin_cvpr2017_fpn}. Specifically, we up-sample the FPN generated multi-scale features to the scale of `$P$2' level by bilinear interpolation, which is 1/4 resolution of the original image. Each feature map is followed by a 1$\times$1 256-d convolutional layer for aligning to the same semantic space, then these feature maps are fused together to generate high-resolution features. 

\noindent\textbf{Global Semantic Feature.} As shown in Figure~\ref{fig:sfpn}, we stack four 3$\times$3 256-d convolutional layers after the high-resolution features to generate global semantic feature. Such a design is simple enough, but also can improve the representation ability of the network. In fact, we have tried some popular enhancement modules for semantic segmentation tasks, such as PPM~\cite{Zhao_cvpr2017_pspnet} and ASPP~\cite{Chen_arxiv2017_deeplabv3}~\cite{Chen_eccv2018_deeplabv3plus}, but experiments show that these modules are not helpful to improve human parsing performance. A 1$\times$1 $C$-dimension ($C$ is the category number) convolutional layer is attached to the global semantic feature to predict human part segmentation. 

\noindent\textbf{Multi-scale Features Fusion.} Through the above structure, we can get high-resolution global semantic feature. It is well known that semantic representation can bring performance gains to bounding-box classification and regression~\cite{He_iccv2017_maskrcnn}~\cite{Chen_cvpr2019_htc}. Therefore, we down-sample the global semantic feature to the scales of $P$3$\sim$$P$6, and use element-wise sum to fuse them with the same scale FPN features. The generated new features are called global semantic enhanced multi-scale features, and denoted as $S$2$\sim$$S$6. We follow the Proposals Separation Sampling~\cite{Yang_cvpr2019_parsingrcnn} strategy that the $S$2$\sim$$S$6 level features are adopted for extracting region features for BBox branch and only $S$2 level is used for Parsing branch.

\subsection{Parsing Re-Scoring Network}

Parsing Re-Scoring Network (PRSN) aims to predict accurate mIoU score for each instance parsing map, and can be flexibly integrated into the Parsing branch.

\noindent\textbf{Concise and Lightweight Design.} PRSN follows the concise and lightweight design, which will not bring too much computation cost to model training and inference. PRSN receives two inputs, one is the $N$$\times$512$\times$32$\times$32 dimension parsing feature map, the other is the $N$$\times$$C$$\times$128$\times$128 dimension segmentation probability map ($N$ is the number of RoIs, $C$ is the category number). A max pooling layer with stride = 4 and kernel = 4 is adopted to make the probability map has the same spatial scale with parsing feature map. The down-sampled probability map and parsing feature map are concatenated together, then followed by two 3$\times$3 128-d convolutional layers. A final global average pooling layer, two 256-d fully connected layers, and MSE loss to regress the mIoU between the predicted instance parsing map and ground-truth.

\noindent\textbf{IoU-aware Ground-truth.} We define the mIoU between the predicted instance parsing map and matched ground-truth as regression target for PRSN. The common Parsing branch can output the segmentation probability map of each human instance, and calculate the loss with the segmentation ground-truth through a cross entropy function. Therefore, the mIoU between them can be calculated directly in the existing framework. It is worth noting that since the instance parsing ground-truth depends on the predicted region of Bbox branch, there is some deviation from the true location of human instance. However, we find that this deviation does not affect the effect of the predicting parsing score, so we do not make corrections to this deviation.

\subsection{Training and Inference}

\begin{figure*}[t]
\begin{center}
\includegraphics[width=0.70\linewidth]{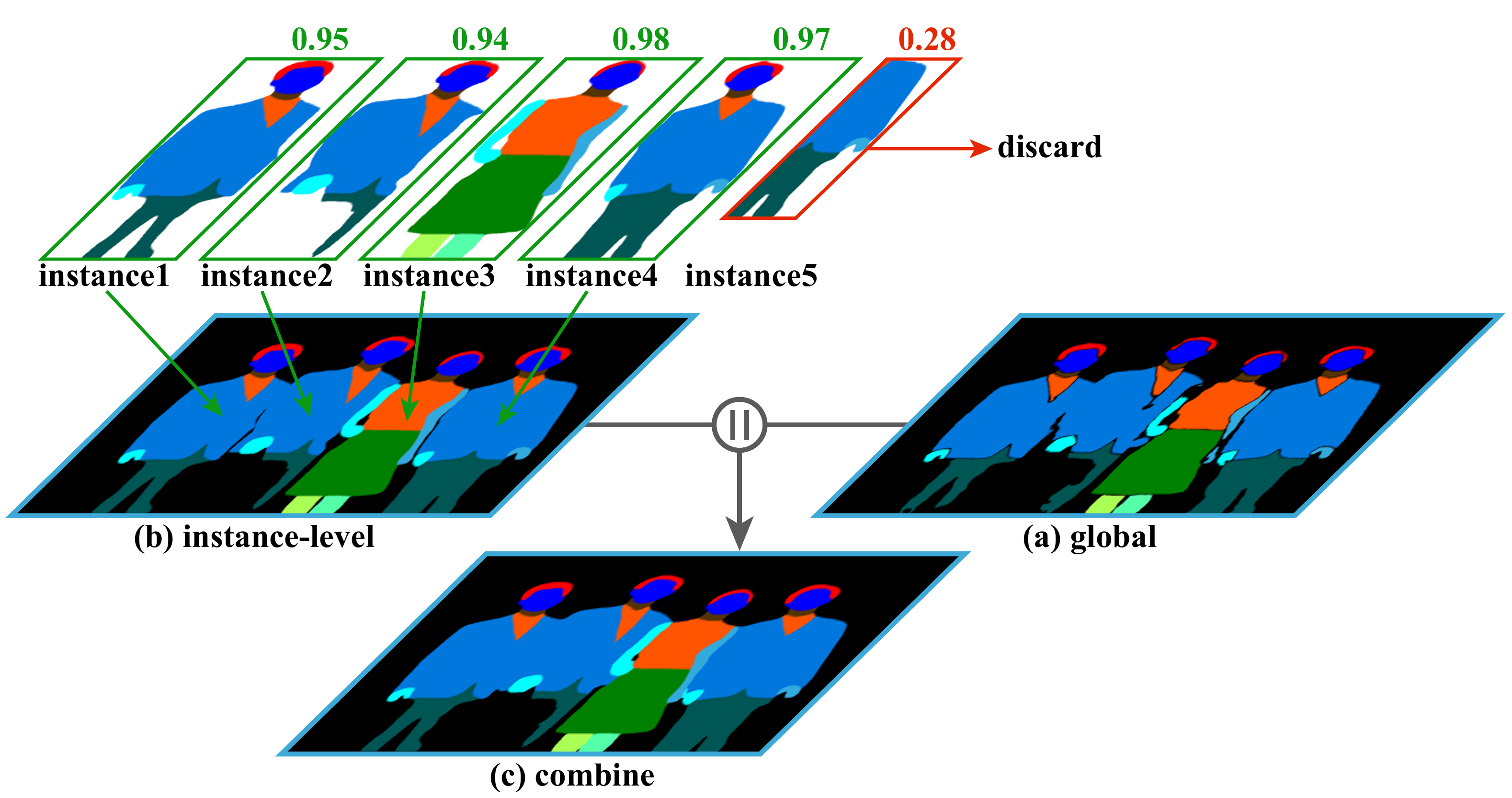}
\end{center}
\caption{Combination strategy for generating semantic segmentation results. ` $\parallel$ ' symbol represents element-wise OR operation.}
\label{fig:seg_infer}
\end{figure*}

As we introduce new supervision into RP R-CNN, there are some changes in the training and inference phases compared with the common methods~\cite{Yang_cvpr2019_parsingrcnn}~\cite{Qin_bmvc2019_unified}.

\noindent\textbf{Training.} There are three losses for global human parts segmentation and Parsing branch: ${\cal{L}}_{\text{sem}}$ (segmentation loss), ${\cal{L}}_{\text{par}}$ (parsing loss), ${\cal{L}}_{\text{res}}$ (re-scoring loss). The segmentation loss and parsing loss are computed as a per-pixel cross entropy loss between the predicted segmentation and the ground-truth labels. We use the MSE loss as re-scoring loss. We have observed that the losses from three tasks have different scales and normalization policies. Simply adding them degrades the overall performance. This can be corrected by a simple loss re-weighting strategy. Considering the losses of the detection sub-network, the whole network loss ${\cal{L}}$ can be written as:

\begin{equation} \label{loss_whole}
\begin{aligned}
   {\cal{L}} = {\cal{L}}_{\text{rpn}} + {\cal{L}}_{\text{bbox}} +  \lambda_\text{p}  {\cal{L}}_{\text{par}}  +  \lambda_\text{s}  {\cal{L}}_{\text{sem}} +  \lambda_\text{r}  {\cal{L}}_{\text{res}}.
\end{aligned}
\end{equation}
The ${\cal{L}}_{\text{rpn}}$ and ${\cal{L}}_{\text{bbox}}$ are losses of RPN and BBox branch, each of which is composed of classification loss and box regression loss. By tuning $\lambda_\text{p}$, $\lambda_\text{s}$ and $\lambda_\text{r}$, it is possible to make the network converge to optimal performance.

\begin{table*}[t]
\centering
\small
\tabcolsep 0.04in 
\scalebox{0.94}{
\begin{tabular}{c|c|cccc}
Dataset & Method & mIoU  &  AP$^\text{p}_\text{50}$ & AP$^\text{p}_\text{vol}$  & PCP$_\text{50}$                     \\
  \hline  
 \multirow{3}{*}{CIHP}  & Parsing R-CNN~\cite{Yang_cvpr2019_parsingrcnn} & 56.3 & 63.7 & 53.9 & 60.1    \\
                                    & Parsing R-CNN (our impl.) & 56.2 & 64.6 & 54.2 & 60.9  \\ 
                                    \cline{2-6}
                                    &  $\Delta$     &  \emph{-0.1} &  \emph{+0.9} &  \emph{+0.4} &  \emph{+0.8}    \\      
  \hline  
 \multirow{3}{*}{MHP-v2}  & Parsing R-CNN~\cite{Yang_cvpr2019_parsingrcnn} & 36.2 & 24.5 & 39.5 & 37.2    \\
                                         & Parsing R-CNN (our impl.) & 35.5 & 26.6 & 40.3 & 37.9    \\
                                         \cline{2-6}
                                         &  $\Delta$ &  \emph{-0.7} &  \emph{+2.1} &  \emph{+0.8} &  \emph{+0.7}    \\                                                
\end{tabular}
}
  \caption{Results of Parsing R-CNN~\cite{Yang_cvpr2019_parsingrcnn} on the CIHP and MHP-v2 datasets. `our impl.' denotes our implementation of Parsing R-CNN, which uses GN~\cite{Wu_eccv2018_gn} in Parsing branch to stabilize the training.}
  \label{tab:baseline}
\end{table*}

\noindent\textbf{Inference.} For network inference, we select top 100 candidate bounding-boxes per image from the human detection results. These candidates are fed into Parsing branch to predict instance parsing map and mIoU score. However, the mIoU score is only trained by positive samples, which leads to the lack of the ability to suppress negative samples. So we fuse mIoU score ${\cal{S}}_{\text{iou}}$ and classification score ${\cal{S}}_{\text{cls}}$ to generate the final parsing score ${\cal{S}}_{\text{parsing}} =  \sqrt{{\cal{S}}_{\text{cls}} * {\cal{S}}_{\text{iou}}}$. In addition, we also find that the global human parts segmentation results are complementary to the Parsing branch result, the former has higher recall for each foreground, and the latter has better details. Thus, when generating semantic segmentation results, we adopt a new combination strategy, as shown in Figure~\ref{fig:seg_infer}. We filter out the low quality results based on the ${\cal{S}}_{\text{parsing}}$, and then generate instance-level human parts segmentation (b) $\emph{instance-level}$. The instance-level (b) was and global human parts segmentation (a) $\emph{global}$ do element-wise OR operation to get the final result (c) $\emph{combine}$. It is worth noting that if the results of (a) and (b) are different at the same pixel, and both of them are predicted as non-background category, we directly adopt the results of (b). This is because the (b) has a more accurate perception of the human parts inside each instance.

\section{Experiments}
 
In this section, we describe experiments on multiple human parsing of RP R-CNN. All experiments are conducted on the CIHP~\cite{Gong_eccv2018_pgn} and MHP-v2~\cite{Zhao_mm2018_mhpv2} datasets. We follow the Parsing R-CNN evaluation protocols. Using mean intersection over union (mIoU)~\cite{Long_cvpr2015_fcn} to evaluate the human part segmentation. And using average precision based on part (AP$^\text{p}$)~\cite{Zhao_mm2018_mhpv2} as instance evaluation metric(s).

\subsection{Implementation Details}

\begin{table*}[t]
\centering
\tabcolsep 0.08in 
\scalebox{0.94}{
\begin{tabular}{c|c|cccc}
 $\lambda_\text{p}$ & AP$^\text{bbox}$ & mIoU  &  AP$^\text{p}_\text{50}$ & AP$^\text{p}_\text{vol}$  & PCP$_\text{50}$                     \\
  \hline  
 0.0  &  \textcolor{gray}{69.1} & -- & -- & -- & --    \\
  \hline 
 0.5  &  \textcolor{gray}{67.7} & 55.9 & 64.4 & 53.7 & 59.9    \\
 1.0  &  \textcolor{gray}{68.5} & 55.9 & 63.9 & 53.8 & 60.6    \\
 \textbf{2.0}  &  \textcolor{gray}{68.3} & \textbf{56.2} & \textbf{64.6} & \textbf{54.3} & \textbf{60.9}    \\
 3.0  &  \textcolor{gray}{67.8} & 55.9 & \textbf{64.6} & 54.2 & 60.7    \\    
\end{tabular}
}
  \caption{Weight ($\lambda_\text{p}$) of parsing loss. All models are trained on CIHP \texttt{train} set and evaluated on CIHP \texttt{val} set (with $\lambda_\text{s} = 0.0$ and $\lambda_\text{r} = 0.0$).}
  \label{tab:parsing_loss_weight}
\end{table*}

\begin{table}[t]
\tabcolsep 0.04in 
\begin{minipage}{0.48\linewidth}
\scalebox{0.94}{
\begin{tabular}{c|c|cccc}
 $\lambda_\text{s}$ & AP$^\text{bbox}$ & mIoU  &  AP$^\text{p}_\text{50}$ & AP$^\text{p}_\text{vol}$  & PCP$_\text{50}$                     \\
  \hline  
 0.0  &  \textcolor{gray}{69.1} & \textcolor{gray}{56.2} & \textcolor{gray}{64.6} & \textcolor{gray}{54.3} & \textcolor{gray}{60.9}    \\
  \hline  
 0.5  &  \textcolor{gray}{67.9} & 57.0 & 65.1 & 54.6 & 61.1    \\
 1.0  &  \textcolor{gray}{67.7} & 57.8 & 66.5 & 55.0 & 61.7    \\
 \textbf{2.0}  &  \textcolor{gray}{67.4} & \textbf{58.2} &  \textbf{67.4} &  \textbf{55.5} &  \textbf{62.1}    \\
 3.0  &  \textcolor{gray}{67.1} & 58.0 & 67.4 & 55.3 & 61.8    \\                
 \hline           
 $\Delta$ &   &\emph{+2.0} &  \emph{+2.8} &  \emph{+1.2} &  \emph{+1.2}    \\       
\end{tabular}
}
  \caption{Weight ($\lambda_\text{s}$) of semantic segmentation loss (with $\lambda_\text{r} = 0.0$).}
  \label{tab:semseg_loss_weight}
\end{minipage}
 \hfill
\begin{minipage}{0.48\linewidth}
\scalebox{0.94}{
\begin{tabular}{c|c|cccc}
 $\lambda_\text{s}$ & AP$^\text{bbox}$ & mIoU  &  AP$^\text{p}_\text{50}$ & AP$^\text{p}_\text{vol}$  & PCP$_\text{50}$                     \\
  \hline  
 0.0  &  \textcolor{gray}{69.1} & \textcolor{gray}{56.2} & \textcolor{gray}{64.6} & \textcolor{gray}{54.3} & \textcolor{gray}{60.9}    \\
  \hline  
 0.5  &  \textcolor{gray}{68.2} & 56.3 & 70.1 & 57.4 & 61.1    \\
 \textbf{1.0}  &  \textcolor{gray}{68.3} & \textbf{56.4} &  \textbf{70.3} &  \textbf{57.6} &  \textbf{61.3}    \\
 2.0  &  \textcolor{gray}{68.2} & 56.3 & 70.2 & 57.5 & \textbf{61.3}   \\
 3.0  &  \textcolor{gray}{68.1} & 56.2 &  \textbf{70.3} & 57.5 & 61.1    \\                
 \hline           
 $\Delta$ &   &\emph{+0.2} &  \emph{+5.7} &  \emph{+3.3} &  \emph{+0.4}    \\     
\end{tabular}
}
  \caption{Weight ($\lambda_\text{r}$) of re-scoring loss (with $\lambda_\text{s} = 0.0$).}
  \label{tab:rescore_loss_weight}
\end{minipage}
\end{table}

\begin{table*}[t]
\centering
\small
\tabcolsep 0.08in 
\scalebox{0.94}{
\begin{tabular}{l|ccc}
Inference methods & mIoU  &  Pixel acc. & Mean acc.                   \\
  \hline  
 baseline         & 56.2 & 89.3 & 67.0    \\
  \hline 
 (a) semseg     & 50.2 & 88.0 & 61.3   \\ 
 (b) parsing      & 57.4 & 89.8 & 67.1    \\ 
 \textbf{(c) combine}    & \textbf{58.2} & \textbf{90.2} & \textbf{69.0}   \\
 \hline           
 $\Delta$ &  \emph{+2.0} &  \emph{+0.9} &  \emph{+2.0}    \\
\end{tabular}
}
  \caption{Semantic segmentation results of different inference methods on CIHP dataset. All models are trained on \texttt{train} set and evaluated on \texttt{val} set.}
  \label{tab:semseg_infer}
\end{table*}

\noindent\textbf{Training Setup.} All experiments are based on Pytorch on a server with 8 NVIDIA Titan RTX GPUs. We use 16 batch-size (2 images per GPU) and adopt ResNet50~\cite{He_cvpr2016_resnet} as backbone. The short side of input image is resized randomly sampled from [512, 864] pixels, and the longer side is limited to 1,400 pixels; inference is on a single scale of 800 pixels. Each image has 512 sampled RoIs for Bbox branch and 16 sampled RoIs for Parsing branch. For CIHP dataset, there are 135,000 iterations (about 75 epochs) of the training process, with a learning rate of 0.02 which is decreased by 10 at the 105,000 and 125,000 iteration. For MHP-v2 dataset, the max iteration is half as long as the CIHP dataset with the learning rate change points scaled proportionally.

\noindent\textbf{Parsing R-CNN Re-Implementation.} In order to better illustrate the advantages of RP R-CNN, we have re-implemented Parsing R-CNN according to the original paper~\cite{Yang_cvpr2019_parsingrcnn}. We find that the training of Parsing R-CNN is not very stable. We solve this issue by adding group normalization~\cite{Wu_eccv2018_gn} after each convolutional layer of Parsing branch. As shown in Table~\ref{tab:baseline}, our re-implemented Parsing R-CNN achieves comparable performance with original version both on CIHP and MHP-v2 datasets. Therefore, this work takes our re-implemented Parsing R-CNN as the baseline.

\begin{table*}[t]
\centering
\small
\tabcolsep 0.04in 
\scalebox{0.94}{
\begin{tabular}{c|cc|cccc}
Methods &  GSE-FPN & PRSN & mIoU  &  AP$^\text{p}_\text{50}$ & AP$^\text{p}_\text{vol}$  & PCP$_\text{50}$                     \\
  \hline  
 \multirow{4}{*}{RP R-CNN}   &                  &   & 56.2 & 64.6 & 54.3 & 60.9    \\
                                            \cline{4-7}
                                            & \checkmark & & ${58.2}_{\color{red}(+2.0)}$ & ${67.4}_{\color{red}(+2.8)}$ & ${55.5}_{\color{red}(+1.2)}$ & ${62.1}_{\color{red}(+1.2)}$  \\ 
                                            &  &\checkmark & ${56.4}_{\color{red}(+0.2)}$ & ${70.3}_{\color{red}(+5.7)}$ & ${57.6}_{\color{red}(+3.3)}$ & ${61.3}_{\color{red}(+0.4)}$  \\ 
                                            &\checkmark&\checkmark   & ${58.2}_{\color{red}(+2.0)}$ & ${71.6}_{\color{red}(+7.0)}$ & ${58.3}_{\color{red}(+4.0)}$ & ${62.2}_{\color{red}(+1.3)}$   \\
\end{tabular}
}
  \caption{Ablations of RP R-CNN on CIHP dataset. All models are trained on \texttt{train} set and evaluated on \texttt{val} set.}
  \label{tab:ablation_cihp}
\end{table*}

\begin{table*}[t]
\centering
\small
\tabcolsep 0.04in 
\scalebox{0.94}{
\begin{tabular}{c|cc|cccc}
Methods &  GSE-FPN & PRSN & mIoU  &  AP$^\text{p}_\text{50}$ & AP$^\text{p}_\text{vol}$  & PCP$_\text{50}$                     \\
  \hline  
 \multirow{4}{*}{RP R-CNN}   &                  &    & 35.5 & 26.6 & 40.3 & 37.9    \\
                                            \cline{4-7}
                                            & \checkmark & & ${37.3}_{\color{red}(+1.8)}$ & ${28.9}_{\color{red}(+2.3)}$ & ${41.1}_{\color{red}(+0.8)}$ & ${38.9}_{\color{red}(+1.0)}$  \\ 
                                            &  &\checkmark & ${35.7}_{\color{red}(+0.2)}$ & ${39.6}_{\color{red}(+13.0)}$ & ${44.9}_{\color{red}(+4.6)}$ & ${38.2}_{\color{red}(+0.3)}$  \\ 
                                            &\checkmark&\checkmark   & ${37.3}_{\color{red}(+1.8)}$ & ${40.5}_{\color{red}(+13.9)}$ & ${45.2}_{\color{red}(+4.9)}$ & ${39.2}_{\color{red}(+1.3)}$   \\
\end{tabular}
}
  \caption{Ablations of RP R-CNN on MHP-v2 dataset. All models are trained on \texttt{train} set and evaluated on \texttt{val} set.}
  \label{tab:ablation_mhp}
\end{table*}

\subsection{Ablation Studies}

In this sub-section, we assess the effects of different settings and components on RP R-CNN by details ablation studies. 

\noindent\textbf{Loss Weights.} To combine our GSE-FPN with PRSN in Parsing R-CNN, we need to determine how to train a single, unified network. Previous studies demonstrate that multi-task training is often challenging and can lead to degraded results~\cite{Kendall_cvpr2018_mtl}. We also observe that adding the losses of all tasks directly will not give the best results. But grid searching three hyper-parameters ($\lambda_\text{p}$, $\lambda_\text{s}$ and $\lambda_\text{r}$) is very inefficient, so we first determine $\lambda_\text{p}$, and then determine $\lambda_\text{s}$ and $\lambda_\text{r}$ separately to improve efficiency. As shown in Table~\ref{tab:parsing_loss_weight}, we find that the network performance is the best when $\lambda_\text{p} = 2.0$. We consider the group normalization layer makes the network convergence more stable, so that a proper large loss weight will bring higher accuracy. Table~\ref{tab:semseg_loss_weight} shows that $\lambda_\text{s} = 2.0$ is the proper loss weight for semantic segmentation. Although increasing the weight of global human parts segmentation loss will slightly reduce the accuracy of human detection, the overall performance is improved. Table~\ref{tab:rescore_loss_weight} shows that the re-scoring task is not sensitive to the loss weight, and its loss scale is smaller than other losses, so it has no significant impact on the optimization of other tasks in multi-task training. To sum up, we choose $\lambda_\text{p} = 2.0$, $\lambda_\text{s} = 2.0$ and $\lambda_\text{r} = 1.0$ as the loss weights of Eqn.(\ref{loss_whole}).

\noindent\textbf{Inference Methods.} The combination inference method proposed in Figure~\ref{fig:seg_infer} utilizes the complementarity of global human parts segmentation and Parsing branch results, and we give the detailed results in Table~\ref{tab:semseg_infer}. The performance of using (a) $\emph{semseg}$ or (b) $\emph{parsing}$ alone is poor. The combination method (c) $\emph{combine}$ can significantly improve the metrics of semantic segmentation, and outperforms the baseline by 2 points mIoU.

\begin{table*}[t]
\centering
\small
\tabcolsep 0.04in 
\scalebox{0.75}{
\begin{tabular}{c|l|l|c|cccc}
Dataset &      Methods &  Backbones & Epochs & mIoU  &  AP$^\text{p}_\text{50}$ & AP$^\text{p}_\text{vol}$  & PCP$_\text{50}$                     \\
  \hline  
 \multirow{18}{*}{CIHP~\cite{Gong_eccv2018_pgn}} & \multicolumn{2}{c|}{Bottom-Up} &\\
                                       \cline{2-8}       
                                      & PGN$^\dagger$~\cite{Gong_eccv2018_pgn} & ResNet101 & $\sim$80 & 55.8 & 34.0 & 39.0 & 61.0    \\             
                                      & DeepLab v3+~\cite{Chen_eccv2018_deeplabv3plus} & Xception & 100 & 58.9 & -- & -- & --    \\
                                      & Graphonomy~\cite{Gong_cvpr2019_graphonomy} & Xception & 100 & 58.6 & -- & -- & --      \\
                                      & GPM~\cite{He_aaai2020_grapyml} & Xception & 100 & 60.3 & -- & -- & --      \\
                                      & Grapy-ML~\cite{He_aaai2020_grapyml} & Xception & 200 & 60.6 & -- & -- & --      \\
                                       \cline{2-8}                       
                                       & \multicolumn{2}{c|}{Two-Stage Top-Down} &\\
                                        \cline{2-8}       
                                      & M-CE2P~\cite{Ruan_aaai2019_ce2p} & ResNet101 & 150 & 59.5 & -- & -- & --      \\
                                      & BraidNet~\cite{Liu_mm2019_braidnet} & ResNet101 & 150 & 60.6 & -- & -- & --      \\
                                      & SemaTree~\cite{Ji_arxiv2019_sematree} & ResNet101 & 200 & 60.9 & -- & -- & --       \\
                                       \cline{2-8}                       
                                       & \multicolumn{2}{c|}{One-Stage Top-Down} &\\
                                       \cline{2-8}      
                                      & Parsing R-CNN~\cite{Yang_cvpr2019_parsingrcnn} & ResNet50 & 75 & 56.3 & 63.7 & 53.9 & 60.1    \\
                                      & Parsing R-CNN$^\dagger$~\cite{Yang_cvpr2019_parsingrcnn} & ResNeXt101 & 75 & 61.1 & 71.2 & 56.5 & 67.7    \\
                                      & Parsing R-CNN (our impl.) & ResNet50 & 75 & 56.2 & 64.6 & 54.3 & 60.9    \\
                                      & Unified~\cite{Qin_bmvc2019_unified} & ResNet101 & $\sim$37 & 55.2 & 51.0 & 48.0 & --      \\
                                      & RP R-CNN (ours) & ResNet50 & 75 & 58.2 & 71.6 & 58.3 & 62.2    \\
                                      & RP R-CNN (ours)$^*$ & ResNet50 & 150 & 60.2 & 74.1 & 59.5 & 64.9    \\
                                      & \textbf{RP R-CNN (ours)}$^*$$^\dagger$ & \textbf{ResNet50} & 150 & \textbf{61.8} & \textbf{77.2} & \textbf{61.2} & \textbf{70.5}    \\
  \hline                                        
\multirow{12}{*}{MHP-v2~\cite{Zhao_mm2018_mhpv2}}& \multicolumn{2}{c|}{Bottom-Up} &\\
                                       \cline{2-8}      
                                      & MH-Parser~\cite{Li_arxiv2017_mhparser} & ResNet101 & -- & -- & 17.9 & 36.0 & 26.9    \\
                                      & NAN~\cite{Zhao_mm2018_mhpv2} & -- & $\sim$80 & -- & 25.1 & 41.7 & 32.2    \\
                                      \cline{2-8}                       
                                       & \multicolumn{2}{c|}{Two-Stage Top-Down} &\\
                                        \cline{2-8}       
                                      & M-CE2P~\cite{Ruan_aaai2019_ce2p} & ResNet101 & 150 & \textbf{41.1} & 34.5 & 42.7 & \textbf{43.8}      \\
                                      & SemaTree~\cite{Ji_arxiv2019_sematree} & ResNet101 & 200 & -- & 34.4 & 42.5 & 43.5       \\
                                      \cline{2-8}                        
                                       & \multicolumn{2}{c|}{One-Stage Top-Down} &\\
                                       \cline{2-8}  
                                      & Mask R-CNN~\cite{He_iccv2017_maskrcnn} & ResNet50 & --  & -- & 14.9 & 33.8 & 25.1    \\
                                      & Parsing R-CNN~\cite{Yang_cvpr2019_parsingrcnn} & ResNet50 & 75 & 36.2 & 24.5 & 39.5 & 37.2    \\
                                      & Parsing R-CNN (our impl.) & ResNet50 & 75 & 35.5 & 26.6 & 40.3 & 37.9    \\
                                      & RP R-CNN (ours) & ResNet50 & 75 & 37.3 & 40.5 & 45.2 & 39.2    \\
                                      & \textbf{RP R-CNN (ours)}$^*$ & \textbf{ResNet50} & 150 & 38.6 & \textbf{45.3} & \textbf{46.8} & \textbf{43.8}    \\
\end{tabular}
}
  \caption{Multiple human parsing on the CIHP and MHP-v2 datasets. $^*$ denotes longer learning schedule. $^\dagger$ denotes using test-time augmentation.}
  \label{tab:sota}
\end{table*}

\noindent\textbf{Ablations on RP R-CNN.} In Table~\ref{tab:ablation_cihp}, we perform the additional ablations of RP R-CNN on CIHP dataset. We observe that GSE-FPN is very helpful to the global human parts segmentation, which yields 2.0 points mIoU improvement. In addition, the global semantic feature also improves the instance metrics, AP$^\text{p}_\text{50}$, AP$^\text{p}_\text{vol}$ and PCP$_\text{50}$ increase 2.8, 1.2 and 1.2 points respectively. With PRSN, the improvements of instance metrics are very significant, AP$^\text{p}_\text{50}$ improves 5.7 points, and AP$^\text{p}_\text{vol}$ improves 3.3 points. With GSE-FPN and PRSN, our proposed RP R-CNN achieves 58.2 mIoU and 71.6 AP$^\text{p}_\text{50}$ on CIHP. The additional ablations on MHP-v2 dataset is shown in Table~\ref{tab:ablation_mhp}. Through GSE-FPN and PRSN, the performance of human parsing is also significant improved. Particularly, AP$^\text{p}_\text{50}$ and AP$^\text{p}_\text{vol}$ are raised considerably by PRSN, 13.0 points and 4.6 points, respectively.

\begin{figure*}[t]
\begin{center}
	\includegraphics[width=0.42\linewidth]{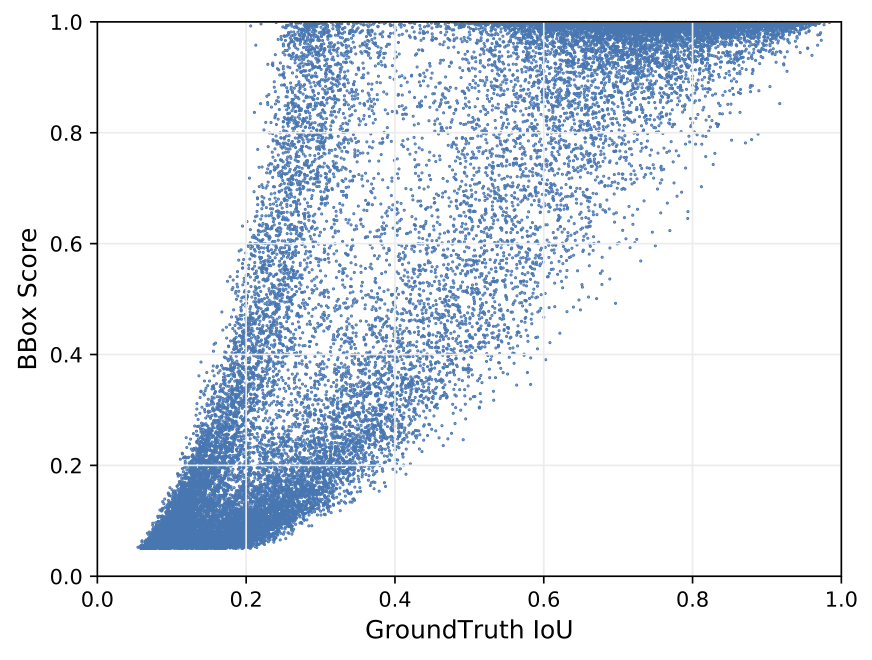}
	\includegraphics[width=0.42\linewidth]{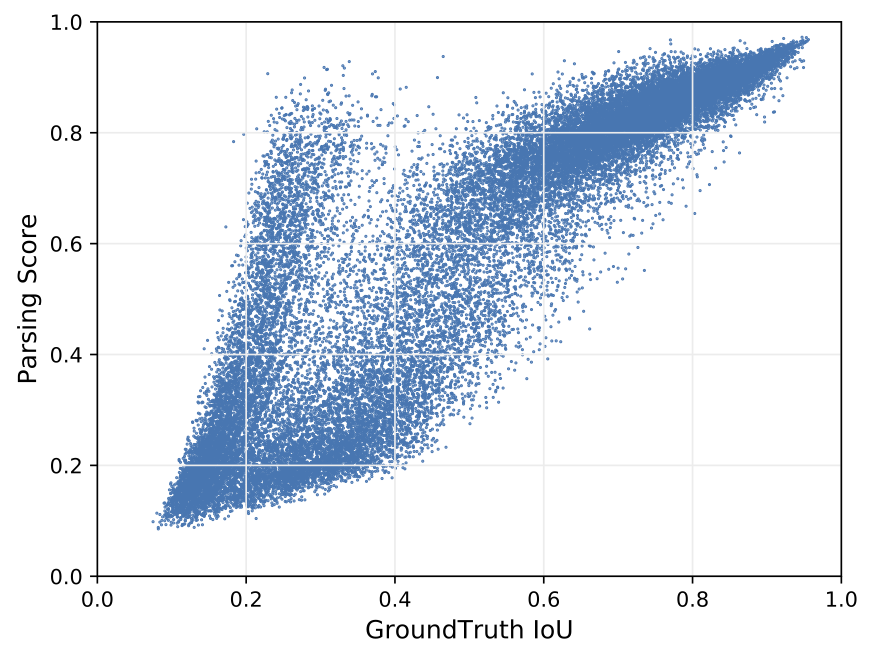}
\end{center}
\caption{Comparisons of ground-truth IoU vs. bbox score (\textbf{Left}) and ground-truth IoU vs. parsing score (\textbf{Right}) on CIHP dataset. All models are trained on \texttt{train} set and evaluated on \texttt{val} set.}
\label{fig:correlation}
\end{figure*}

\subsection{Comparison with State-of-the-Arts}

We evaluate RP R-CNN on the CIHP and MHP-v2 datasets and compare the results to state-of-the-art including bottom-up and one-stage/two-stage top-down methods, shown in Table~\ref{tab:sota}. On CIHP dataset, our proposed RP R-CNN achieves 58.2 mIoU and 71.6 AP$^\text{p}_\text{50}$, which surpasses Parsing R-CNN~\cite{Yang_cvpr2019_parsingrcnn} in all respects. Compared with PGN~\cite{Gong_eccv2018_pgn}, the performance advantage of RP R-CNN is huge, and AP$^\text{p}_\text{50}$ is even 37.6 points ahead. With longer learning schedule (150 epochs), RP R-CNN achieves compared performance with one-stage top-down methods,  {\em e.g.} M-CE2P and BraidNet. Even though we have adopted a lighter backbone (ResNet50 vs. ResNet101). Finally, using test-time augmentation, RP R-CNN with RestNet50 achieves state-of-the-art performance on CIHP.

On MHP-v2 dataset, RP R-CNN achieves excellent performance. We can observe that our RP R-CNN outperforms Parsing R-CNN consistently for all the evaluation metrics. And compared with M-CE2P, RP R-CNN yields about 10.8 point AP$^\text{p}_\text{50}$ and 4.1 points AP$^\text{p}_\text{vol}$ improvements. With RestNet50 backbone, it gives new state-of-the-art of 45.3 AP$^\text{p}_\text{50}$, 46.8 AP$^\text{p}_\text{vol}$ and 43.8 PCP$_\text{50}$.

\subsection{Analysis and Discussion}

\noindent\textbf{Effect of Parsing Re-Scoring Network.} Figure~\ref{fig:correlation} shows that the correlation between mIoU of the predicted parsing map with the matched ground-truth and the bbox/parsing score. As shown, the parsing score has better correlation with the ground-truth, especially for high-quality parsing map. However, it is difficult to score low-quality parsing map, which is the source of some false positive detections. Therefore, the evaluation of low-quality prediction is still a problem to be solved.

\begin{figure*}[t]
\begin{center}
\includegraphics[width=0.8\linewidth]{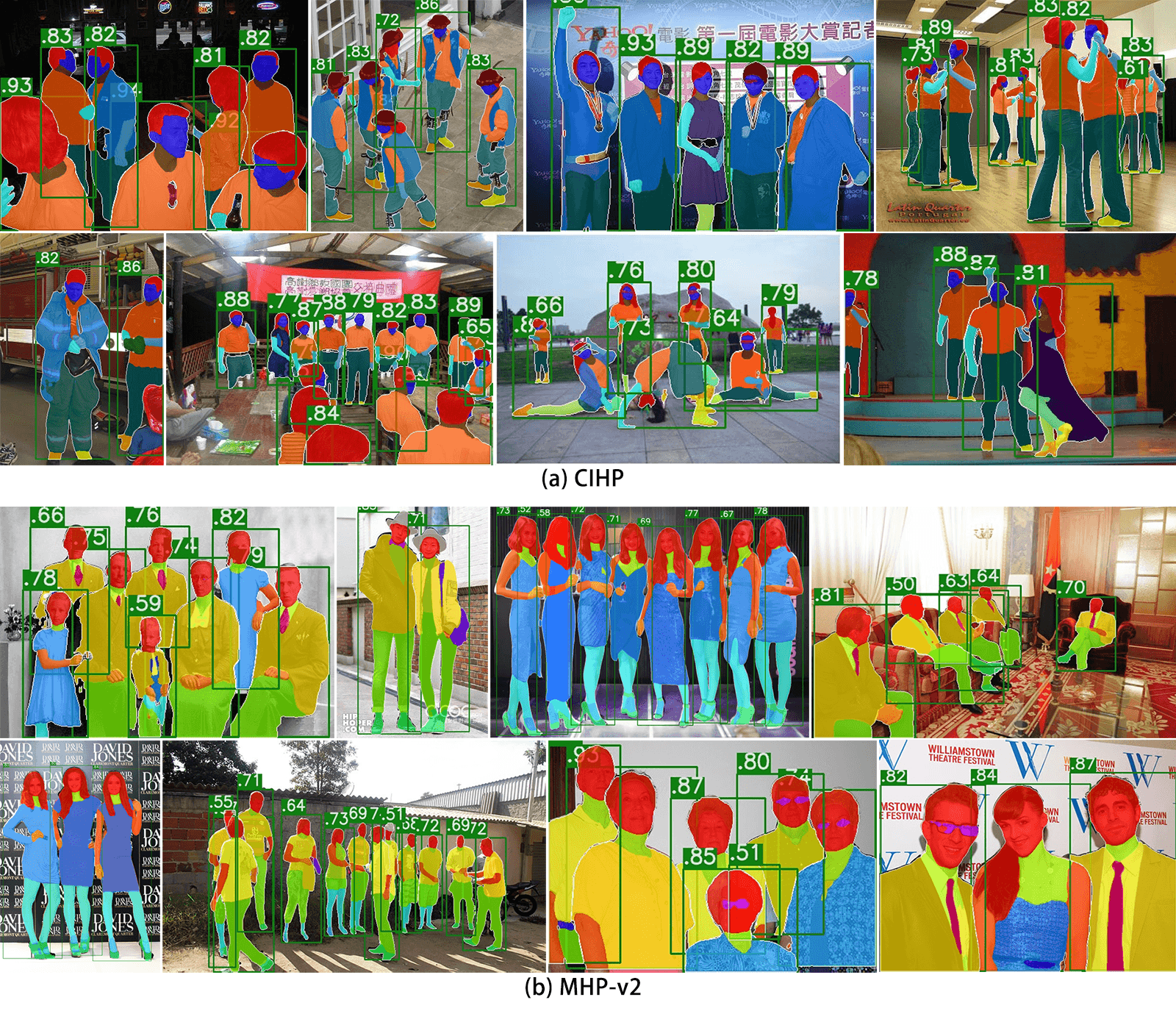}
\end{center}
\caption{Qualitative results of RP R-CNN on the CIHP and MHP-v2 datasets.}
\label{fig:qresults}
\end{figure*}

\noindent\textbf{Qualitative results.} We visualize multiple human parsing results of RP R-CNN in Figure~\ref{fig:qresults}. We can observe that RP R-CNN has a good applicability to dense crowds and occlusions. In addition, the parsing score predicted by RP R-CNN reflects the quality of the parsing map.

\section{Conclusions}

In this paper, we proposed a novel Renovating Parsing R-CNN (RP R-CNN) model for solving the issue of missing global semantic information and inaccurate scoring of parsing result in top-down multiple human parsing. By explicitly introducing global semantic enhanced multi-scale features and learning the mIoU between instance parsing map and matched ground-truth, our RP R-CNN outperforms previous state-of-the-art methods consistently for all the evaluation metrics. In addition, we also adopt a new combination strategy, which improves the results of multiple human parsing by global semantic segmentation and instance-level semantic segmentation. We hope our effective approach will serve as a cornerstone and help the future research in multiple human parsing.

\clearpage
%
%
\bibliographystyle{splncs04}
\bibliography{egbib}
\end{document}